\documentclass{bmvc2k}

\usepackage{times}
\usepackage{epsfig}
\usepackage{graphicx}
\usepackage{amsmath}
\usepackage{amssymb}
\usepackage{enumitem}
\usepackage{dsfont}
\usepackage{booktabs} % table (toprule, bottomrule)
\usepackage{multirow, makecell} % table
\usepackage{kotex}
\definecolor{blue_js}{RGB}{0,26,102}
\title{HairFIT: Pose-Invariant Hairstyle Transfer \\via Flow-based Hair Alignment and Semantic-Region-Aware Inpainting}
\addauthor{Chaeyeon Chung}{cy_chung@kaist.ac.kr}{$\ast$1,2}
\addauthor{Taewoo Kim}{specia1ktu@kaist.ac.kr}{$\ast$1,2}
\addauthor{Hyelin Nam}{nhl0208@cau.ac.kr}{$\ast$3}
\addauthor{Seunghwan Choi}{shadow2496@kaist.ac.kr}{1}
\addauthor{Gyojung Gu}{gyojung.gu@kaist.ac.kr}{1}
\addauthor{Sunghyun Park}{psh01087@kaist.ac.kr}{1,2}
\addauthor{Jaegul Choo}{jchoo@kaist.ac.kr}{1}

\addinstitution{
KAIST AI\\
South Korea
}
\addinstitution{
Kakao Enterprise\\
South Korea
}
\addinstitution{
Chung-Ang University\\
South Korea
}

\runninghead{Chung et el.}{HairFIT, Pose-Invariant Hairstyle Transfer}

\def\eg{\emph{e.g}\bmvaOneDot, }

\def\ie{\emph{i.e}\bmvaOneDot, }

\newcommand{\ourmodel}{HairFIT\xspace}
\newcommand{\maskblock}{SIM estimator\xspace}
\newcommand{\hairsynthesis}{ALIAS\xspace}

\newcommand{\REV}[1]{\textcolor{black}{#1}}

\newcommand{\Rthree}[1]{\textcolor{black}{#1}}
\newcommand{\Rfour}[1]{\textcolor{black}{#1}}
\newcommand{\Rfive}[1]{\textcolor{black}{#1}}
\newcommand{\RtwoRthree}[1]{\textcolor{black}{#1}}

\begin{document}

\maketitle
\vspace{-0.4cm}
\begin{abstract}

%Hairstyle transfer is challenging, since it requires preservation of delicate hair features, such as texture, shape, and color, while considering head pose difference and occluded regions at the same time. 
%쓸거면 더 자세히 쓰거나 빼거나
Hairstyle transfer is the task of modifying a source hairstyle to a target one.
Although recent hairstyle transfer models can reflect the delicate features of hairstyles, they still have two major limitations.
First, the existing methods fail to transfer hairstyles when a source and a target image have different poses (\eg viewing direction or face size), which is prevalent in the real world.
Also, the previous models generate unrealistic images when there is a non-trivial amount of regions in the source image occluded by its original hair.
When modifying long hair to short hair, shoulders or backgrounds occluded by the long hair need to be inpainted.
To address these issues, we propose a novel framework for pose-invariant hairstyle transfer, \ourmodel.
Our model consists of two stages: 1) flow-based hair alignment and 2) hair synthesis.
In the hair alignment stage, we leverage a keypoint-based optical flow estimator to align a target hairstyle with a source pose.
Then, we generate a final hairstyle-transferred image in the hair synthesis stage based on Semantic-region-aware Inpainting Mask (SIM) estimator.
Our SIM estimator divides the occluded regions in the source image into different semantic regions to reflect their distinct features during the inpainting.
To demonstrate the effectiveness of our model, we conduct quantitative and qualitative evaluations using multi-view datasets, K-hairstyle and VoxCeleb.
The results indicate that \ourmodel achieves a state-of-the-art performance by successfully transferring hairstyles between images of different poses, which have never been achieved before.
\end{abstract}

\section{Introduction}
\label{sec:intro}
Recently, hairstyle has been considered as a way to express one’s own identity. 
Responding to the increasing demands, various approaches have tackled virtual hairstyle transfer, a task of modifying one's hairstyle to a target one, based on generative adversarial networks~\cite{goodfellow2014generative}.
%has increased as it presents how hairstyles will look on them.
The existing methods~\cite{tan2020michigan, saha2021loho} transfer the target hairstyle to a source image while successfully preserving delicate features of the target hairstyle. 
However, despite their convincing performances, hairstyle transfer still poses several challenges. 

%First, a model is required to reflect delicate texture and color of the target hairstyle, which are not always globally consistent.
%Even one hairstyle can contain multiple texture and color. 
%Also, they are sensitive to surrounding features such as lighting. 
% First -> 이 부분이 우리 논문에서 tackling 하는게 아니므로 이렇게 다루는 것에대해 다시 생각해봐야함.
% \sh{Second}, occlusions in the source image due to its original hair are required to be inpainted. 
% For instance, if a model attempts to transfer a long hair of the source image to the short hair, shoulders and backgrounds which are originally occluded needs to be newly generated to apply the short hair.
% Lastly, a model needs to consider the differences in a head pose or face size of the source and the target image. 
% Since the change of a viewing angle or a scale significantly deforms the shape of a hairstyle, the model should align the target hair to the source image when they have different views or scales.

First, the existing models are hardly applicable to a source image that has a different pose (\eg viewing direction or face size) from a target image, which is prevalent in the real world. 
The first two rows of \REV{Fig.~\ref{fig:qualitative}} present examples of the pose difference between the source and the target image.
%Since the change of pose deforms the hairstyle, the target hairstyle should have a similar pose with the source image.
Since the same hairstyle appears different when its pose changes, the target hairstyle should have a similar pose with the source image.
%view (\eg a head pose) or scale (\eg a face size) 
However, the previous approaches fail to align the target hair with the source pose when their poses are significantly different.

Additionally, the existing methods poorly generate regions originally occluded in the source image.
When the source image contains the regions that are occluded by its original hair, the occlusions need to be newly generated during the transfer.
For instance, the \REV{first row of Fig.~\ref{fig:sim_comparison}} shows that the occluded shoulders and backgrounds should be properly inpainted when transferring the shorter hair.
Moreover, the occlusions become even larger when the source and the target image have a significant pose difference.
Although the existing models leverage the external inpainting networks, it is still challenging for a model to fill in the occlusions containing multiple semantic regions.

To address these issues, we propose a novel framework, Hairstyle transfer via Flow-based hair alignment and semantic-region-aware InpainTing (\ourmodel), which successfully performs hairstyle transfer regardless of a pose difference between a source and a target image.
\ourmodel consists of two parts: 1) flow-based hair alignment and 2) hair synthesis.
We first align the target hair with the source pose based on the estimated optical flow between the source and the target.
Then, we apply the aligned target hair to the source image to synthesize the final output.
The hair synthesis module inpaints in the occluded regions of the source image and refines the aligned target hair leveraging ALIgnment-Aware Segment (ALIAS) generator~\cite{choi2021viton}.
We newly present a Semantic-region-aware Inpainting Mask (SIM) estimator to support realistic occlusion inpainting. By predicting separate semantic regions, the estimator helps our generator to reflect distinctive features of diverse semantic regions (\eg backgrounds, clothes, and forehead) during inpainting.

% 따라서 우리는 motion transfer를 기반으로한 GAN-based hairstyle translation framework를 제시한다.
% multi view hairstyle dataset인 khairstyle dataset을 기반으로 진행
% -> 우리가 해결 1) alignment, inpainting 모듈을 따로했다

We conduct quantitative and qualitative evaluations using K-hairstyle and VoxCeleb which contain images of different poses. 
The results indicate that our model achieves a state-of-the-art performance compared to the existing methods. 
Our model outperforms other models especially when the source and the target image have a significantly different pose.
% we achieved the state-of-the-art performance compared to the existing models
% real world dataset에서도 좋은 성능
% Contribution
% achieving the state-of-the-art pose-invariant hairstyle translation performance based on motion transfer and GAN.
% SAM block

% \begin{figure*}[t!]
%     \centering
%     \includegraphics[width=\linewidth]{images/challenge_eg.png}
%     \vspace{-0.8cm}
%     \caption{Hairstyle transfer results on challenging cases. The first row shows the case of a significant pose difference between a source and a target, and the second row presents the case of occlusion in a source image due to its long hair. Despite these challenges, our model generates the most realistic outputs.}
%     \vspace{-0.2cm}
%     \label{fig:challenges} 
% \end{figure*} 

Our contributions are summarized as follows:
\begin{itemize}
    \item This is the first work that proposes a pose-invariant hairstyle transfer framework by adopting the optical flow estimation for hair alignment.
    \vspace{-0.2cm}
     \item We present a Semantic-region-aware Inpainting Mask (SIM) estimator which supports high-quality occlusion inpainting by allowing hair synthesis module to reflect distinctive features of each segmented inpainting region.
    \vspace{-0.2cm}
    \item \ourmodel achieves the state-of-the-art performance in both quantitative and qualitative evaluations with two multi-view datasets, K-hairstyle and VoxCeleb.
\end{itemize}

%----------------------------------------------------------------------
% 지금까지의 도메인 발전 과정 + 우리 연구와의 접점
% 우리 연구를 이해하기 위해 알면 좋은 것들 소개

\section{Related Work}
\subsection{Conditional Image Generation} 
% conditional gan generates ~ 정의 그대로 -> (ACGAN)
% 특히 사람 얼굴 이미지에 있어서 특정 feature (face shape, hairstyle, etc) 를 바꾸는 work이 있다 -> MaskGAN ~ Deep Plastic Surgery 
% 이런거는 보통 mask나 sketch 를 condition input으로 사용해서 하는데, 이 input을 manually edit 해야한다
% 우리는 user interation없이 automaitc하게 hair style을 바꾸는 모델을 제안한다
Conditional image generation is the task of synthesizing images based on the given conditions such as category labels~\cite{brock2018large, odena2017conditional}, text~\cite{reed2016generative,zhang2018stackgan++}, and images~\cite{isola2017image, park2019SPADE}.
Recent studies on image-conditioned generation have proposed various approaches to modify specific features (\eg face shape, hairstyle, etc.) on facial images~\cite{lee2020maskgan, jo2019sc, portenier2018faceshop, yang2020deep}.
In particular, MaskGAN~\cite{lee2020maskgan} enables facial image manipulation based on a user-edited semantic mask as a conditional input.
Also, SC-FEGAN~\cite{jo2019sc}, FaceShop~\cite{portenier2018faceshop}, and Deep Plastic Surgery~\cite{yang2020deep} allow modification of facial attributes based on human-drawn sketches.
However, the existing methods require a non-trivial amount of user interaction to obtain the desired conditional inputs to change specific target attributes.
In this paper, we propose a framework that transfers a hairstyle without any user-modified conditional input or user interaction.

% Conditional generative adversarial networks (cGANs) generate the images utilizing inputs such as category labels~\cite{brock2018large, odena2017conditional}, text~\cite{reed2016generative,zhang2018stackgan++}, and images~\cite{isola2017image, park2019SPADE} as conditioning information.
% In the recent few years, image-conditioned generation have actively developed, starting with pix2pix~\cite{isola2017image}.
% In particular, a few studies can produce high-quality facial images, which is difficult to generate.
% For instance, MaskGAN~\cite{lee2020maskgan} generates an image based on semantic masks, allowing users to manipulate the image by modifying semantic masks. SC-FEGAN~\cite{jo2019sc}, FaceShop~\cite{portenier2018faceshop}, and Deep Plastic Surgery~\cite{yang2020deep} modify facial attributes based on human-drawn sketches.
% However, these methods are unsuitable for users who want to synthesize specific attributes of target images as they are.
% In this paper, we propose a method that synthesizes hairstyles like a target image without any manipulation, which is intuitive and convenient.

\subsection{Hairstyle Transfer}
Hairstyle transfer aims to modify a hairstyle of a source image to a target one while preserving other features of the source.
Along with the researches on facial attribute modification~\cite{portenier2018faceshop,jo2019sc,yang2020deep}, hairstyle transfer has also been actively tackled in the recent few years~\cite{tan2020michigan, saha2021loho}.
MichiGAN~\cite{tan2020michigan} successfully transfers a hairstyle reflecting its delicate features using disentangled hair attributes and conditional synthesis modules of each attribute.
In addition, LOHO~\cite{saha2021loho} generates realistic hairstyle-transferred images by decomposing hairstyle features and optimizing latent space to inpaint missing hair structure details.
However, these works only tackle the cases where poses of a source and a target image are similar. Therefore, they fail to generate realistic images when the source and the target have significantly different viewing directions or face sizes.
Unlike the existing approaches, our model successfully transfers a hairstyle even when the source and the target image have a significant pose difference.

\subsection{Optical Flow Estimation}
Optical flow represents apparent pixel-level motion patterns between two images using a vector field.
There are several approaches based on neural networks that directly predict an optical flow map between two consecutive frames of a video~\cite{dosovitskiy2015flownet,ilg2017flownet,hui2018liteflownet} and warp the raw pixels of a frame to generate the unseen frame~\cite{liu2017video, liang2017dual, gao2019disentangling, park2021vid}.
The optical flow estimation is also leveraged to transfer motions from another video by learning the motion difference between a source and a target frame~\cite{siarohin2019animating, siarohin2019first}.
To achieve pose-invariant hairstyle transfer, we utilize the optical flow to deform the target hairstyle according to the pose of the source image. 
To the best of our knowledge, this is the first work that applies optical flow estimation to hairstyle transfer.

\section{Method}

\begin{figure*}[t!]
    \centering
    \includegraphics[width=\linewidth]{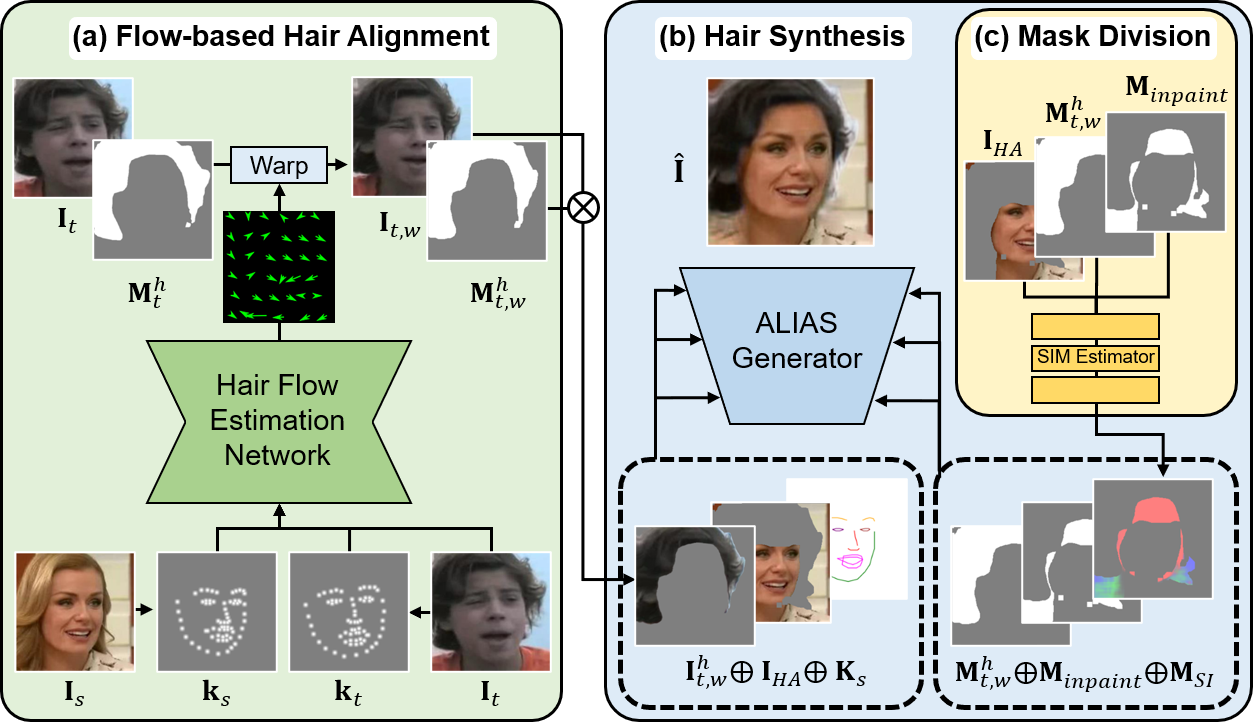}
    %\vspace{-0.7cm}
    \caption{Overall architecture of \ourmodel. Our model consists of two modules: (a) flow-based hair alignment module and (b) hair synthesis module. (a) Hair alignment module aligns the hair from a target image $\mathbf{I}_t$ with the pose of a source image $\mathbf{I}_s$ based on facial keypoints $\mathbf{k}_s$, $\mathbf{k}_t$ extracted from each image. In (b) hair synthesis stage, \hairsynthesis generator synthesizes the final image $\mathbf{\hat{I}}$ based on (c) mask division with SIM estimator.
    % 최종본
    }
    \vspace{-0.5cm}
    \label{fig:framework}
\end{figure*}

%\jg{method overview 를 설명하고, 이때 이런 데이터셋이 필요한데, unpaired 는 없으니까, multiview dataset을 활용해서 해당 task 를 풀려고한다. 전체적인 구조를 설명하고 각각 dataset 과 proposed 한 모듈들이 왜 필요한지 말하기. 각 모듈의 중요도에 따라서 비중 구성하기. overview에서도 강조할 부분 강조하기. 큰 변화없이 차용한 부분들은 양을 줄이기도 해야하고. 기존 연구는 간략히 설명하되 큰 그림은 제시. 새롭게 우리가 제안하는 모듈에 집중해서 서술. VITON, Region Norm, SPADE 설명 구조 참고.}

\subsection{Overview}
\label{sec:method:overview}
We employ a two-stage approach with two separate modules: 1) flow-based hair alignment module and 2) hair synthesis module.
% \ourmodel consists of two modules:
When given a source image and a target image containing the target hairstyle, the flow-based hair alignment module first warps the target image and its hair segmentation mask to align the target hairstyle with the source face. 
Here, we utilize the optical flow predicted by our hair flow estimation network.
% We also warp a target hair segmentation mask with the same optical flow to obtain a hair mask of the warped target image. 
Then, the hair synthesis module generates the final image based on the aligned target hair and the source image via semantic-region-aware inpainting and hair refinement.
Fig.~\ref{fig:framework} describes the overall architecture of \ourmodel.

We train our model with multi-view datasets, which contain diverse views for each individual.
In particular, we sample two images (\ie a source and a target image), which have different views with the same identity and hairstyle.
Then, our model is required to reconstruct the source image by transferring the hairstyle of the target image to the source. Here, the source image works as the ground truth the model aims to generate.
In this manner, the flow-based hair alignment module learns to align the target hair with the source pose according to the pose difference, and the hair synthesis module learns to reconstruct the original source image based on the aligned target hairstyle. % 어떤 pair 들이 input 으로 들어가는지를 적기

Ideally, when given a source and a target image with different poses, identities, and hairstyles, the ground truth image with the source pose and identity and the target hairstyle can provide \ourmodel with direct supervision.
In this case, the ground truth image can guide the model to align and apply the target hairstyle to the source face.
However, a hairstyle dataset containing such images does not exist since it is expensive and time-consuming to collect them.
% Therefore, we train our model with multi-view datasets.
Instead, we utilize the multi-view datasets for training our model.
Although our model uses only image pairs of the same identity and hairstyle during the training, our model successfully transfers the hairstyle of the target image that has a different identity and hairstyle from the source.

\subsection{Flow-based Hair Alignment}

The flow-based hair alignment module aims to align the pose (\ie view and scale) of a target image with a source image.

As described in Fig.~\ref{fig:framework} (a), our hair flow estimation network estimates a dense optical flow map that represents a pose difference between the source image $\mathbf{I}_s$ and the target image $\mathbf{I}_t$ with the source keypoints $\mathbf{k}_s$, target keypoints $\mathbf{k}_t$, and $\mathbf{I}_t$ given as inputs
% $\mathbf{k}_s$ and $\mathbf{k}_t$ indicate the facial keypoints $\mathbf{k} \in \mathbb{R}^{N_k \times 2}$ extracted from $\mathbf{I}_s$ and $\mathbf{I}_t$, where $N_k$ indicates the number of keypoints.
\REV{($\mathbf{I} \in \mathbb{R}^{3 \times H \times W}$ and $\mathbf{k} \in \mathbb{R}^{N_k \times 2}$, where $H$, $W$, and $N_k$ denote the image height, width, and the number of keypoints, respectively)}.
Then, we warp $\mathbf{I}_t$ as well as its hair segmentation mask $\mathbf{M}_t^h$ according to the estimated optical flow.
As a result, we gain the warped target hairstyle in $\mathbf{I}_{t,w}$ and its hair mask $\mathbf{M}_{t,w}^h$ aligned with the source face, where $h$ indicates hair and $w$ means the image or mask is warped.

\noindent \textbf{Hair flow estimation network.}
Hair flow estimation network predicts the dense optical flow with the given $\mathbf{k}_s, \mathbf{k}_t,$ and $\mathbf{I}_t$. 
To obtain the coarse pose difference, the network first converts $\mathbf{k}_s$ and $\mathbf{k}_t$ into a Gaussian keypoint heatmap $\mathbf{H} \in \mathbb{R}^{N_k \times H \times W}$, respectively. % %based on the keypoint difference and estimates local regions .
Then, we obtain the keypoint heatmap difference $\mathbf{\hat{H}}$, calculated as $\mathbf{H}_t - \mathbf{H}_s$.
With $\mathbf{\hat{H}}$ and $\mathbf{I}_t$ deformed by the keypoint difference $\mathbf{\hat{k}}$, which is ${\mathbf{k}}_t - {\mathbf{k}}_s$, the network predicts the $N_k$-channel mask $\mathbf{M} \in \mathbb{R}^{N_k \times H \times W}$ and $\mathcal{F}_{ref}$.
$\mathbf{M}$ contains the estimated local regions to apply each channel of $\mathbf{\hat{k}}$ to.
%$\mathbf{M}$ informs how to compose the coarse optical flow map reflecting the keypoint difference of each feature map. 
Also, $\mathcal{F}_{ref}$ indicates a refinement flow map that reflects additional detailed optical flow.
% Additionally, the network leverages $\mathbf{I}_t$ warped by $\mathbf{\hat{k}}$, which is calculated as ${\mathbf{k}}_t - {\mathbf{k}}_s$, in the estimation.
Finally, we obtain the dense optical flow $\mathcal{F} \in \mathbb{R}^{2 \times H \times W}$ for the entire image by adding the coarse optical flow and refinement flow map as $\mathcal{F} = \sum_{i=1}^{N_k} \rho(\mathbf{\hat{k}}^i) \otimes \mathbf{M}^i + \mathcal{F}_{ref}$.
Here, $\rho(\cdot)$ repeats the input tensor by $H \times W$ times and $\otimes$ denotes element-wise multiplication.

More details of the network architectures are described in supplementary materials. 
For training, we use the reconstruction loss $\mathcal{L}_{rec}$, which is calculated as $\mathbb{E}[\rVert \mathbf{I}_{t,w} - \mathbf{I}_s\rVert_{1}]$.
For the implementation, we referred to the existing keypoint-based optical flow estimation networks~\cite{siarohin2019animating, siarohin2019first}.

\noindent \textbf{Facial keypoints.}
Our hair flow estimation network utilizes keypoints to predict the flow map between the source and the target image.
Here, the predicted optical flow needs to reflect the pixel-wise pose difference effectively, thus the keypoints should represent the overall view and scale of an image~\cite{siarohin2019animating, wang2021one}.
%Moreover, as mentioned in Section~\ref{sec:method:overview}, our model needs to transfer a hairstyle between images of different identities at test time, unlike the training setting where only the same identity pairs exist.
In this regard, we utilize the facial keypoints, which consistently represent a person's view (\eg head pose) or scale (\eg face size), regardless of his/her identity.
Facial keypoints are simple but robust pose representations that guide the network to capture the key differences in poses.
Since the datasets do not include ground truth facial keypoints, we extract 68 facial keypoints from each image using the pre-trained keypoint detector~\cite{bulat2017far}.

\begin{figure*}[!h]
    \centering
    \includegraphics[width=\linewidth]{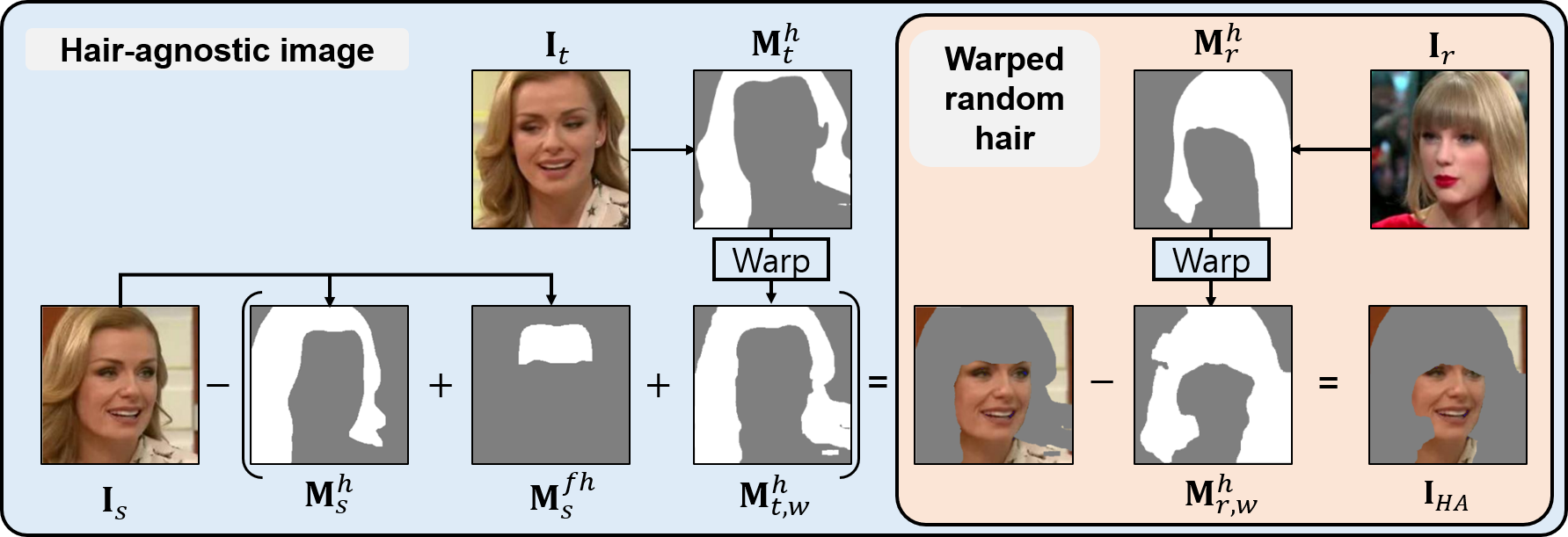}
    \vspace{-0.5cm}
    \caption{Process to obtain the hair-agnostic image $\mathbf{I}_{HA}$ for hair synthesis. First, we remove the source hair region $\mathbf{M}_s^{h}$, the forehead region $\mathbf{M}_s^{fh}$, and the warped target hair region $\mathbf{M}_{t,w}^{h}$ from $\mathbf{I}_s$. We additionally remove a random warped hair region $\mathbf{M}_{r,w}^{h}$ during training to obtain $\mathbf{I}_{HA}$. $\mathbf{M}_{t,w}^{h}$ and $\mathbf{M}_{r,w}^{h}$ are warped by our flow-based hair alignment module.
    } 
    \vspace{-0.2cm}     
    \label{fig:hairagnostic}
\end{figure*}

\subsection{Hair Synthesis}
\label{subsec:hair}
% Keep image 어떻게 얻는지 설명 
% SEIN 에서 어떻게 normalizatoin and denormalization 되는지 구조 추가 
The goal of the hair synthesis module is to transfer the target hairstyle to the source image using the aligned target hairstyle obtained from the previous stage.
To achieve this, we utilize \hairsynthesis generator~\cite{choi2021viton} with a newly-proposed hair-agnostic image and a Semantic-region-aware Inpainting Mask (SIM) estimator.

\noindent \textbf{Hair-agnostic image.}
%As mentioned in Section~\ref{sec:method:overview}, our model is trained with image pairs of the same identity and hairstyle since we leverage multi-view datasets. 
At test time, our model needs to transfer hairstyle between the images of different identities and hairstyles, unlike the training phase. 
Therefore, the difference between the target hairstyle and the source hairstyle during the test is much larger than during the training.
To minimize the difference, we obtain the hair-agnostic image $\mathbf{I}_{HA}$ with the hair region of the source image $\mathbf{I}_s$ removed as presented in Fig.~\ref{fig:hairagnostic}.
To be specific, \REV{we exclude the region of the source hair mask $\mathbf{M}_s^{h}$ and the forehead mask $\mathbf{M}_s^{fh}$ from $\mathbf{I}_s$.
The warped target hair mask $\mathbf{M}_{t,w}^{h}$ is also subtracted from $\mathbf{I}_s$.
During the training, $\mathbf{M}_{t,w}^{h}$ and $\mathbf{M}_{s}^{h}$ have only a slight difference since we use image pairs of the same hairstyle.
Therefore, we sample a random image $\mathbf{I}_r$ from the training set and warp its hair mask $\mathbf{M}_{r}^{h}$ to fit the source pose, resulting in $\mathbf{M}_{r,w}^{h}$. 
Then, the region of $\mathbf{M}_{r,w}^{h}$ is removed from the hair-agnostic image.
This successfully mimics the inference situations by providing the model with the chance to fill in larger occluded regions.
In this regard, the proposed $\mathbf{I}_{HA}$ enhances our hair synthesis generator's generalization ability at test time.}

\noindent \textbf{\maskblock.}
As previously mentioned, the hairstyle transfer model is required to inpaint the occlusions in the source image and refine the aligned target hair.
To address the challenges, we propose a SIM estimator, which separates the occluded regions into different semantic areas.
With $\mathbf{I}_{HA}, \mathbf{M}_{t,w}^h,$ and $\mathbf{M}_{inpaint}$ as inputs, \REV{\maskblock predicts a semantic-region-aware inpainting mask $\mathbf{M}_{SI} \in \{0,1\}^{4 \times H \times W}$ by dividing $\mathbf{M}_{inpaint}\in \{0,1\}^{1 \times H \times W}$ into face, clothes, background, and unknown. Here, $\mathbf{M}_{inpaint}$ is calculated by subtracting $\mathbf{M}_{t,w}^h$ from the occluded region of $\mathbf{I}_{HA}$.} The fifth column in Fig.~\ref{fig:sim_comparison} presents examples of $\mathbf{M}_{SI}$.
The separated mask effectively guides our \hairsynthesis generator to inpaint each semantic region by reflecting its distinctive features, leading to high-quality image generation.

\noindent \REV{\textbf{\hairsynthesis generator.} As described in Fig.~\ref{fig:framework} (b), the inputs  $\mathbf{I}_{t,w}^h \oplus \mathbf{I}_{HA} \oplus \mathbf{K}_s$ and $\mathbf{M}_{t,w}^h \oplus \mathbf{M}_{inpaint} \oplus \mathbf{M}_{SI}$ are injected into each layer of \hairsynthesis generator, where $\mathbf{K}_s \in \mathbb{R}^{3 \times H \times W}$ is an RGB-rendered source keypoint image.
In each layer, the features are normalized separately based on the inpainting mask $\mathbf{M}_{inpaint}$ in a similar manner to general region normalization~\cite{yu2020region, choi2021viton}.  %source and the warped target hair 
The normalized features are then modulated by $\mathbf{\beta}$ and $\mathbf{\gamma}$, which are predicted based on $\mathbf{M}_{t,w}^{h} \oplus \mathbf{M}_{inpaint} \oplus \mathbf{M}_{SI}$.
As a semantic guidance map of inpainting, $\mathbf{M}_{SI}$ allows the modulation parameters $\mathbf{\beta}$ and $\mathbf{\gamma}$ to reflect distinctive semantic features of each divided region, increasing the overall quality of occlusion inpainting of the output. 
The hair synthesis module generates the final output after a series of \hairsynthesis residual blocks with up-sampling layers.
Additional details of \hairsynthesis generator are provided in our supplementary material.
}

\noindent \textbf{Losses.}
We train the \hairsynthesis generator with the conditional adversarial loss $\mathcal{L}_{cGAN}$, the feature matching loss $\mathcal{L}_{FM}$, the perceptual loss $\mathcal{L}_{percept}$, and the hairstyle loss $\mathcal{L}_{style}$, referring to SPADE~\cite{park2019semantic}, pix2pixHD~\cite{wang2018high}, and LOHO~\cite{saha2021loho}.  % 마지막 h perecpt 수정
Since the training pairs have the same identity and hairstyle, the losses can be computed between a synthesized image and the ground truth (source) image, working as direct supervision.
We use the Hinge loss as the adversarial loss~\cite{zhang2019self}. 
The $\mathcal{L}_{style}$ is computed between Gram matrix~\cite{gatys2016image} feature maps of the generated hair image and the source hair image to support detailed hairstyle refinement. 
The generated hair image is extracted from the final output using the pre-trained hair segmentation model~\cite{gong2019graphonomy}. 
\REV{Additionally, we train \maskblock with SIM loss $\mathcal{L}_{SIM}$.
$\mathcal{L}_{SIM}$ is calculated using the binary cross-entropy loss between the estimated $\mathbf{M}_{SI}$ and the ground truth segmentation masks obtained by the pre-trained face parsing model~\cite{yu2018bisenet}.}
% Each of the loss functions are followed as:
% \begin{equation}
% %\begin{split}
%     L_{cGAN} = \mathbb{E}[\log(D(M_{div},I_s))] + \mathbb{E}[1-\log(D(M_{div},\hat{I}))]    
% %\end{split}
% \end{equation}
% \vspace{-0.5cm}
% \begin{equation}
% %\begin{split}
%     L_{FM} = \mathbb{E}\sum_{i=1}^T \frac{1}{E_{i}}[\rVert D^i(M_{div}, I_s) - D^i(M_{div}, \hat{I}))\rVert_{1}]
% %\end{split}
% \end{equation}
% \vspace{-0.2cm}
% \begin{equation}
%     L_{percept} = \mathbb{E}\sum_{i=1}^V \frac{1}{R_{i}}[\rVert F^i(I_s) - F^i(\hat{I})\rVert_{1}]
% \end{equation}
% \vspace{-0.2cm}
% \begin{equation}
%     L_{style} = \mathbb{E}\sum_{i=1}^V \frac{1}{R_{i}}[\rVert G^i(F^i(M_s^h \odot I_s)) - G^i(F^i(M_g^h \odot \hat{I}))\rVert_{2}]
% \end{equation}
% \begin{equation} % check the equation BCE
% %\begin{split}
%     L_{SI} = -\mathbb{E}[GT_{SI}\log(M_{SI}) + (1-GT_{SI})\log(1-M_{SI})]  % 수식 수정해야함. 시그마 붙이기    
% %\end{split}
% \end{equation}
The total loss of the hair synthesis module is as follows:

\begin{equation}
    %\vspace{0.1cm}
    \mathcal{L}_{total} = \mathcal{L}_{cGAN} + \lambda_{FM} \mathcal{L}_{FM} +
    \lambda_{percept} \mathcal{L}_{percept} + \lambda_{style} \mathcal{L}_{style} + \lambda_{SIM} \mathcal{L}_{SIM},
    \vspace{0.4cm}
\end{equation} 
where we set both $\lambda_{FM}$ and $\lambda_{percept}$ as 10, $\lambda_{style}$ as 50, and $\lambda_{SIM}$ as 100. 
More details of the losses are described in the supplementary material.

\vspace{-0.3cm}
\section{Experiments}

To demonstrate the effectiveness of \ourmodel, we conduct quantitative and qualitative evaluations compared to baseline models, using two multi-view datasets. 
We also conduct an ablation study to present the effect of key components in our model.

\vspace{-0.3cm}
% khairstyle
\subsection{Experimental Setup}
\noindent \textbf{Dataset.}
As mentioned in Section~\ref{sec:method:overview}, we leverage multi-view datasets for the experiments.
First, we utilize K-hairstyle~\cite{kim2021k} which contains 500,000 high-resolution hairstyle images. The dataset consists of multi-view images including more than 6,400 identities. The viewing angle of images ranges from 0 to 360 degrees. Also, each image in the dataset has a hair segmentation mask as well as various hairstyle attributes.
Since our goal is to transfer a hairstyle, we excluded images whose hairstyle is significantly occluded. Also, we removed images whose face is extremely rotated, such as images of a person facing the side or the back, since we cannot extract facial keypoints from them.
Accordingly, our training set consists of randomly sampled 60,584 images with 5,407 identities, and the test set includes 6,717 images with 611 identities.
While the maximum image resolution is 4,032$\times$3,024, we resized the images into 256$\times$256 and 512$\times$512 in our experiments.
%In particular, we only use images of facing angles ranging from -30 to 30, when facing front is considered as 0. % angle check. % it's seem to be weird we call our networks as view-invariant... 

% Voxceleb --> multiview dataset, 원래, train, test 장수, identity.
Additionally, we use VoxCeleb~\cite{nagrani2017vox} which consists of more than 100,000 utterance videos with 1,251 identities. 
For the training data, we randomly sampled 16,847 videos, each of which contains 30 frames on average. 
The frames of a video can be considered as multi-view images of a single identity.
The test set includes randomly sampled 2,209 videos. For the experiments, we resized the images into 256$\times$256.

For both datasets, the training set consists of image pairs that have the same identity and hairstyle, while the test set contains image pairs that have different identities and hairstyles. We utilize one image of a pair as a source image and the other as a target image.

% 다양한 identity의 멀티뷰 데이터셋인~
% + vox1) 1251 celeb, 100,000 이상 utterance
% 1) 우리는 리소스의 문제로? (Q. 이유 설명해야?) identity 별로 하나의 발화 선택. 랜덤하게 두 프레임 골라서 ~.
% 2) 결과적으로 n개의 identity의 m개의 이미지 사용~ 

% images in the wild (optional or suppli)

%\subsection{Implementation Detail} %Training Strategy

\noindent \textbf{Evaluation metrics.}
% quantitativequalitative and  (FID, LPIPS, SSIM)
As a quantitative evaluation, we use the fréchet inception distance (FID) score~\cite{heusel2017fid}. The FID score measures how similar the distributions of the synthesized images and the real images are. The lower FID score indicates a higher similarity between the images.

\begin{table}[h!]
    \centering
    \small
    \begin{tabular}{c|cc|c}
    \toprule
    Dataset & \multicolumn{2}{c|}{K-hairstyle} & VoxCeleb \\ \midrule
    Resolution & 256 $\times$ 256 & 512 $\times$ 512 & 256 $\times$ 256 \\
    \midrule
    MichiGAN & 30.52 & 36.12 & 81.14 \\ %\hline
    LOHO & 62.44 & 71.79 & 66.97 \\ %\hline
    \ourmodel(Ours) & \textbf{18.53} & \textbf{19.01} & \textbf{17.66} \\ 
    \bottomrule
    \end{tabular}%
    \vspace{0.2cm}
    \caption{Quantitative comparison with the baselines using K-hairstyle and VoxCeleb. We measure the FID scores.
    \vspace{-0.3cm}
    % FID(256) means that FID is calculated on 256x256 images, and FID(512) is calculated on 512x512 images.
    }
    \label{Table:sota}
\end{table}

\subsection{Comparison to Baselines}
\vspace{0.2cm}

\noindent\textbf{Quantitative evaluations.} First, we compare the FID scores between our model and the baseline models,
MichiGAN~\cite{tan2020michigan} and LOHO~\cite{saha2021loho}. We train both models with the same datasets, K-hairstyle and VoxCeleb, based on their official implementation codes.
We utilize a gated convolution network~\cite{yu2019free} for the inpainting modules in MichiGAN and LOHO, as described in LOHO paper.
As presented in Table~\ref{Table:sota}, our model achieves the \REV{lowest} FID score with a large margin compared to the baseline models.

\begin{table}[!h]
    \centering
    \small
    \begin{tabular}{c|c|c|c}
    \toprule
    Pose difference level & Medium & Difficult & Extremely difficult \\
    % PD range & [2.51, 22.44) & [22.44, 36.58) &  [36.58, 136.06)  \\
    \midrule
    MichiGAN & 31.58 & 34.41 & 40.41 \\ %\hline
    LOHO & 57.80 & 62.34 & 95.89 \\ %\hline
    HairFIT (Ours) & \textbf{21.95} & \textbf{21.76} & \textbf{24.60} \\ 
    \bottomrule
    \end{tabular}%
    \vspace{0.2cm}
    \caption{Quantitative comparison with the baselines with three different levels of pose differences using K-hairstyle. We measure the FID scores.}
    \label{Table:posediff}
    %\vspace{-0.5cm}
\end{table}

For further analysis, we also conduct quantitative comparisons with three different levels of pose difference.
\Rfive{We split our test pairs of K-hairstyle into three categories of 2,000 images, `Medium', `Difficult', and 'Extremely difficult'.
As in  LOHO~\cite{saha2021loho}, we use 68 facial keypoints extracted by the pre-trained keypoint detector~\cite{bulat2017far} to measure the pose distance (PD) between the source and the target keypoints as follows:
$\mathrm{PD} = \frac{1}{68}\sum_{i=1}^{68}\rVert \mathbf{k}_s - \mathbf{k}_t \rVert_{2}$.
The result in Table~\ref{Table:posediff} describes that \ourmodel outperforms the other baselines with a larger margin as the degree of pose difference increases from `Medium' to `Extremely difficult'.
MichiGAN and LOHO show poor hairstyle transfer performance on images with significant pose differences, as they stated in their paper.}

% reconstruction fid, lpips, ssim
% 256
\begin{table}[!b]
    \centering
    \small
    \begin{tabular}{c|cc|cc}
    \toprule
    Dataset & \multicolumn{2}{c|}{K-hairstyle} & \multicolumn{2}{c}{VoxCeleb} \\ 
    \midrule
    Metric & SSIM$_{\uparrow}$ & LPIPS$_{\downarrow}$ & SSIM$_{\uparrow}$ & LPIPS$_{\downarrow}$\\
    \midrule
    MichiGAN & 0.7210 & 0.2432 & 0.7021 & 0.2282 \\ %\hline
    LOHO     & 0.7852 & 0.1452 & 0.6389 & 0.2419 \\ %\hline
    HairFIT(Ours) & \textbf{0.8041} & \textbf{0.0841} & \textbf{0.7147} & \textbf{0.1262}\\ 
    \bottomrule
    \end{tabular}%
    \vspace{0.2cm}
    \caption{Quantitative
    comparison with baselines on hairstyle reconstruction. We measure SSIM and LPIPS using K-hairstyle and VoxCeleb. $\uparrow$ indicates the larger the better and $\downarrow$ means the smaller the better.
    %Quantitative comparison with the baselines on K-hairstyle and VoxCeleb datasets where source and target have different pose prevalently. 
    %We evaluate the SSIM and LPIPS.
    % FID(256) means that FID is calculated on 256x256 images, and FID(512) is calculated on 512x512 images.
    }
    %\vspace{-0.2cm}
    \label{Table:sota_recon}
\end{table}

Additionally, to demonstrate that \ourmodel successfully aligns the target hair with the source pose, we conduct a quantitative evaluation on hairstyle reconstruction. We compare our model to MichiGAN~\cite{tan2020michigan} and LOHO~\cite{saha2021loho}.
Given a source image and a target image which have the same identity and hairstyle but different poses, a model reconstructs the source image by applying the target hairstyle extracted from the target image to the source image.
We measure the structural similarity (SSIM)~\cite{wang2004image} and learned perceptual image patch similarity (LPIPS)~\cite{zhang2018unreasonable} using the same 3,000 image pairs.
The higher the SSIM and the lower the LPIPS, the better.
As described in Table~\ref{Table:sota_recon}, our model achieves superior performance over the baseline models. 

\begin{figure*}[!t]
    \centering
    \includegraphics[width=\linewidth]{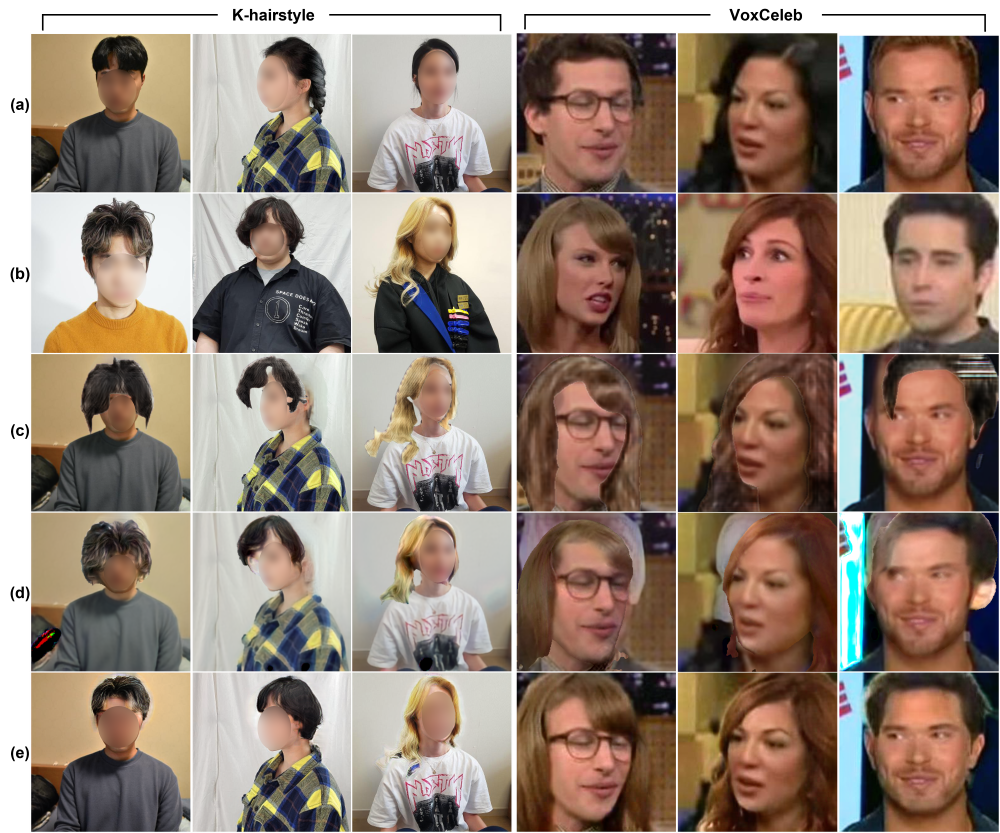}%{images/experiments.png}
    \vspace{-0.2cm}
    \caption{Qualitative comparison with the baselines. (a) indicates the source images, (b) the target images, (c) the results of MichiGAN, (d) LOHO, and (e) ours. Due to the privacy issue, we blur the faces of the images from the K-hairstyle dataset.}
    \vspace{-0.3cm}
    \label{fig:qualitative} 
\end{figure*}

\begin{table}[!b]
    \centering
    \small
    \resizebox{\textwidth}{!}{\begin{tabular}{c|cccccc}
        \toprule
        Method & \begin{tabular}{@{}c@{}} \hairsynthesis \\ only \end{tabular}  & \begin{tabular}{@{}c@{}} + Flow-based \\ hair alignment\end{tabular}  & 
        \begin{tabular}{@{}c@{}} + $\mathbf{M}_{r,w}^{h}$  \\ subtraction \end{tabular} & \begin{tabular}{@{}c@{}} + \maskblock \\ with SIM loss\end{tabular}   & 
        \begin{tabular}{@{}c@{}} + Hairstyle loss \\ (Full)\end{tabular}  \\
        \midrule
        %\midrule
        FID (256) & 25.58 & 22.89 & 19.40 & 19.16 & \textbf{18.53}  \\ % (3.3.1.Hairstyle transfer simulation)
        \bottomrule
    \end{tabular}}
    \vspace{-0.2cm}
    \caption{\RtwoRthree{Ablation study on K-hairstyle dataset. Starting from \hairsynthesis generator, we measure the FID scores by gradually adding each module and loss.}}
    \label{Table:ablation}
\end{table}

\noindent\textbf{Qualitative evaluations.} As presented in Fig.~\ref{fig:qualitative}, qualitative results also present the superiority of \ourmodel for both of the two datasets. While the baseline models generate unrealistic images, our model robustly transfers target hairstyles to the source images even when they have significantly different poses or the source image has occluded regions.

%K-hairstyle and VoxCeleb are multi-view datasets which contain images of  (\ie misaligned images). 

%First, we calculate the FID score of our full model, our model without hair perceptual loss is try-on synthesis network, and our model trained with keypoints estimated by Monkey-Net instead of ground truth keypoints. Table~\ref{Table:ablation} presents that our full model achieves the lowest FID score compared to other settings.

% 간격 동일하게?

% | Settings                 		| FID |
% | only Hair Synthesis network  	| 25.58 |
% | + Flow-based Hair Alignment 	| 22.89 |
% | + Subtraction warped random hair | 19.40 |
% | + SIM Estimator with SIM loss	| 19.16 |
% | + Hairstyle loss (Full)		| 18.53 | 

% \begin{table}[h!]
%     \centering
%     \small
%     \begin{tabular}{c|ccccc}
%         \toprule
%         Method &  w/o $\mathbf{M}_{r,w}^{h}$ subtraction & w/ $\mathbf{M}_{r}^{h}$ subtraction & No \maskblock & Full\\
%         \midrule
%         %\midrule
%         FID(256)  & 22.56 & 20.64 & 19.16 & \textbf{18.53}  \\ % (3.3.1.Hairstyle transfer simulation)
%         \bottomrule
%     \end{tabular}
%     \vspace{0.2cm}
%     \caption{Ablation study on K-hairstyle dataset.}
%     \vspace{-0.4cm}
%     \label{Table:ablation}
% \end{table}

\vspace{-0.2cm}
\subsection{Ablation Study}
\noindent\textbf{Quantitative evaluations.}
\RtwoRthree{We conduct an ablation study to validate the effectiveness of each component in our model. 
Starting from \hairsynthesis generator, we gradually add flow-based hair alignment module, $\mathbf{M}_{r,w}^{h}$ subtraction in a hair-agnostic image $\mathbf{I}_{HA}$, \maskblock with SIM loss, and hairstyle loss, which is our full model.
$\mathbf{M}_{r,w}^{h}$ indicates warped random hair mask in $\mathbf{I}_{HA}$ as described in Section~\ref{subsec:hair}.} \Rfive{The results show that all of our design decisions lead to an improvement of the FID score, successfully addressing both pose differences and occlusion inpainting.}

% First, we explore the influence of hair-agnostic image pre-processing.
% w/o $\mathbf{M}_{r,w}^{h}$ subtraction denotes our model trained with $\mathbf{I}_{HA}$ without subtracting $\mathbf{M}_{r,w}^{h}$.
% Also, w/ $\mathbf{M}_{r}^{h}$ subtraction indicates our model trained with $\mathbf{I}_{HA}$ with subtracting $\mathbf{M}_{r}^{h}$ instead of $\mathbf{M}_{r,w}^{h}$.
% Table~\ref{Table:ablation} shows that our full model using $\mathbf{I}_{HA}$ with $\mathbf{M}_{r,w}^{h}$ subtraction achieves the lowest FID score.
% Furthermore, the FID score of No \maskblock demonstrates that \maskblock efficiently supports our model to inpaint occluded regions separately, leading to more realistic image generation.

\begin{figure*}[!t]
    \centering
    \includegraphics[width=\linewidth]{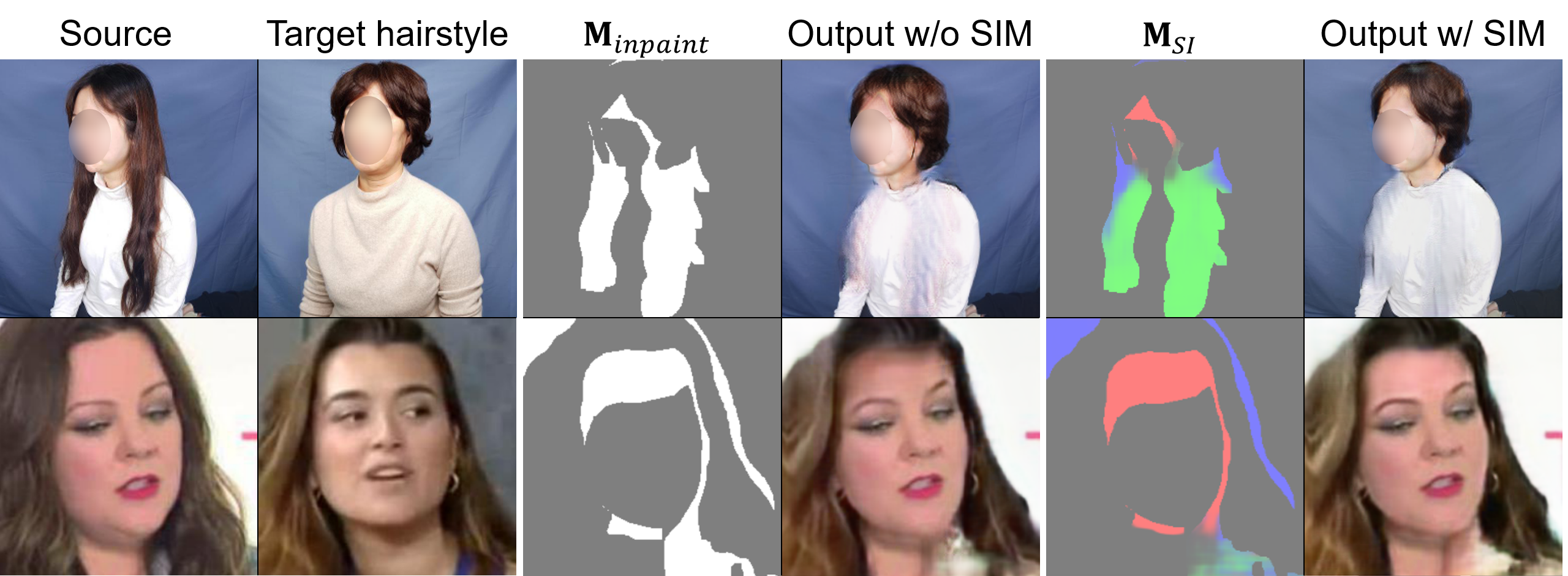}
    \vspace{-0.7cm}
    \caption{Qualitative evaluation on the \maskblock. The first and the second row are examples of K-hairstyle and VoxCeleb, respectively. The red, green, and blue regions of $\mathbf{M}_{SI}$ indicate face, clothes, and background, respectively. Due to the privacy issue, we blur the faces of K-hairstyle images.
}
    %\vspace{-0.5cm}
    \label{fig:sim_comparison} 
\end{figure*} 

% SIM한번더 설명
% SIM 으로 마스크 나눠지는 예시 잘나눠진걸로 한장씩?
\noindent\Rfive{\textbf{Qualitative evaluations.}
Furthermore, we conduct a qualitative evaluation on the effect of \maskblock. 
As mentioned in Section~\ref{subsec:hair}, \maskblock guides our model to effectively inpaint occlusions by reflecting distinctive features of each region.
According to Fig.~\ref{fig:sim_comparison}, outputs of \ourmodel with \maskblock present more realistic inpainting quality compared to \ourmodel without \maskblock. 
In particular, while the output without SIM on the first row of Fig.~\ref{fig:sim_comparison} has blue artifacts on its clothes, the output with SIM does not. 
Moreover, unlike the output without SIM on the second row, the output with SIM has a clean forehead without brown artifacts.}

% \begin{table}[h!]
%     \centering
%     \small
%     \begin{tabular}{cccccC}
%         \toprule
%         Method & FID(256)$_{\downarrow}$ \\
%         \midrule
%         %\midrule
%         No random hair erosion in HS  & 20.81\\ % (3.3.1.Hairstyle transfer simulation)
%         Random unwarped hair erosion in HS  & 19.42 \\ % (3.3.1Hairstyle transfer simulation)
%         No \maskblock in HS  & 19.85 \\ % (3.3.2.\hairsynthesis with \maskblock)
%         No hair style loss in HS  & 18.31  \\ % (3.3.3.hair related loss)
%         Full & \textbf{18.14} \\
%         \bottomrule
%     \end{tabular}
%     \vspace{0.2cm}
%     \caption{Ablation study on K-hairstyle dataset. GT, HA, and HS indicate Ground Truth, Hair Alignment model, and Hair Synthesis module, respectively.} % hair loss 와 관련된 실험은 계속하고 있는데, 결과가 개선되지 않으면 뺄 예정.
%     %\vspace{-0.2cm}
%     \label{Table:ablation}
% \end{table}
\vspace{-0.4cm}
\section{Conclusion}
We propose a two-stage pose-invariant hairstyle transfer model, \ourmodel, which successfully transfers a target hairstyle to a source image when the source and the target have a different pose. 
In our model, a flow-based hair alignment network first aligns the target hairstyle with the source leveraging optical flow estimation. Then, a hair synthesis module generates output via an ALIAS generator with the help of a hair-agnostic image and a SIM estimator.
Our SIM estimator guides the generator to inpaint occlusions in the source image which contain multiple semantic regions.
The quantitative and qualitative results demonstrate the superiority of \ourmodel over the existing methods.
%By introducing hairstyle transfer model that task for the first time, model 
%By dramatically improving FID scores even at the different pose , the bandwidth and ensuring a more immersive experience, 

%We believe that this is an important step towards the future hairstyle tranfer

%his is the first work that proposes a pose-invariant hairstyle transfer by adopting theoptical flow estimation for hair alignment.•  We present a Semantic-region-aware Inpainting Mask (SIM) estimator to divide in-painting regions, which allows hair synthesis module to consider the separated regiondifferently, generating high quality image on regions to be inpainted.•  HairFIT achieves the state-of-the-art performance in both quantitative and qualitativeevaluations with two multi-view datasets, K-hairstyle and VoxCeleb

\section*{Acknowledgement}
This work was supported by the Institute of Information \& communications Technology Planning \& Evaluation (IITP) grant funded by the Korean government (MSIT) (No. 2019-0-00075, Artificial Intelligence Graduate School Program (KAIST) and No. 2020-0-00368, A Neural-Symbolic Model for Knowledge Acquisition and Inference Techniques)), the National Research Foundation of Korea (NRF) grant funded by the Korean government (MSIT) (No. NRF-2019R1A2C4070420), and Kakao Enterprise.

\bibliography{egbib}

\clearpage
\appendix

\noindent \textbf{\huge{\textcolor{blue_js}{Supplementary Material}}}
\section{Model Architecture and Implementation Details}

\subsection{Flow-based Hair Alignment}
As described in Section 3.2 of our paper, the flow-based hair alignment module aligns the target hairstyle with the source pose using a dense optical flow estimated by the hair flow estimation network. We obtain the dense optical flow map $\mathcal{F} \in \mathbb{R}^{2 \times H \times W}$ by combining a coarse keypoint difference and the refinement flow map $\mathcal{F}_{ref} \in \mathbb{R}^{2 \times H \times W}$.

To be specific, our module first converts facial keypoints $\mathbf{k} \in \mathbb{R}^{N_k \times 2}$ into the Gaussian keypoint heatmap $\mathbf{H} \in \mathbb{R}^{N_k \times H \times W}$, where $N_k$ denotes the number of keypoints. Then, we obtain the keypoint heatmap difference $\mathbf{\hat{H}} \in \mathbb{R}^{N_k \times H \times W}$, which is calculated as $\mathbf{H}_t - \mathbf{H}_s$, where $t$ and $s$ indicate a target and a source, respectively.
With $\mathbf{\hat{H}}$ and the warped $\mathbf{I}_t$, the flow estimation network, $FE$, predicts the $N_k$-channel mask $\mathbf{M} \in \mathbb{R}^{N_k \times H \times W}$ and $\mathcal{F}_{ref}$. 
Here, the target image $\mathbf{I}_t$ is warped by the keypoint difference $\mathbf{\hat{k}} \in \mathbb{R}^{N_k \times 2} $, which is calculated as ${\mathbf{k}}_t - {\mathbf{k}}_s$.
Accordingly, $\mathbf{M}$ and $\mathcal{F}_{ref}$ are obtained as follows:
\begin{equation}
    \mathbf{M}, \mathcal{F}_{ref} = FE(\hat{\mathbf{H}} \oplus \mathcal{W}(\mathbf{I}_t,\hat{\mathbf{k}})).
\end{equation}
$\oplus$ and $\mathcal{W}$ indicate concatenation and warping operation, respectively.
$\mathcal{W}(\mathbf{I},\alpha)$ means an image $\mathbf{I}$ is warped by $\alpha$.
The warping operation is implemented using a bilinear sampler.
$FE$ consists of two $1\times1$ convolutional (Conv) blocks, five down blocks, and five up blocks~\cite{siarohin2019animating}.

Finally, we obtain the dense optical flow map $\mathcal{F}$ as
% \begin{equation}
$
    \mathcal{F} = \sum_{i=1}^{N_k} \rho(\mathbf{\hat{k}}^i) \otimes \mathbf{M}^i + \mathcal{F}_{ref}$. 
% \end{equation}
Here, $\rho(\cdot)$ repeats the input tensor by $H \times W$ times and $\otimes$ denotes element-wise multiplication.  
% monkeynet에 dense motion flow 어디를 참고했다 언급
% 3D neural talking head 논문에서 언급한 것: 2D neural taling head 에서 어느정도 가져왔다고 서술
We adopt the Adam optimizer~\cite{kingma2014adam} with $\beta_1 = 0.5$, $\beta_2 = 0.999$ and the learning rate 0.0002. \Rthree{Also, we train the flow-based hair alignment for 240,000 iterations with batch size 8 in case of K-hairstyle dataset and 110,000 iterations with batch size 16 for VoxCeleb.}

\subsection{Hair Synthesis}

\label{subsec:hairsynthesis}
% \vspace{-0.6cm}

\begin{figure*}[h!]
    \centering
    \includegraphics[width=\linewidth]{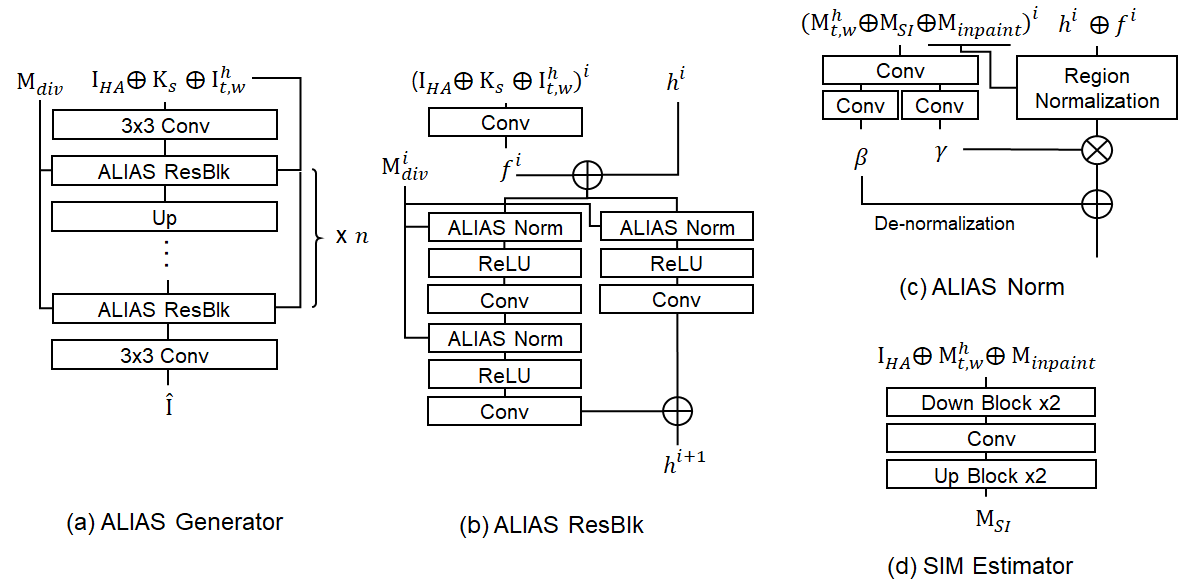}
    \vspace{-0.5cm}
    \caption{Detailed architecture of hair synthesis network.
    }
    \vspace{-0.3cm}
    \label{fig:alias_details} 
\end{figure*}

The detailed architecture of the hair synthesis module is shown in Fig.~\ref{fig:alias_details}. 
Our hair synthesis module synthesizes the aligned target hairstyle with the source image. 
The network needs to refine and apply the warped target hairstyle while preserving the source features such as the faces, clothes, or backgrounds.
Furthermore, the network is also required to inpaint occluded regions in the source image with appropriate face, clothes, or backgrounds.
To achieve this, we utilize ALIgnment-Aware Segmen (ALIAS)~\cite{choi2021viton} generator with Semantic-region-aware Inpainting Mask (SIM) estimator.

\noindent\textbf{ALIAS generator and \Rthree{discriminator}.} As described in Fig.~\ref{fig:alias_details} (a), the generator contains a series of ALIAS residual blocks (ResBlk), along with up-sampling layers.
\Rthree{We use the multi-scale discriminator~\cite{Huang2017adain, choi2021viton}.}

\noindent\textbf{ALIAS ResBlk.} As presented in Fig.~\ref{fig:alias_details} (b), each ALIAS ResBlk consists of three sets of ALIAS normalization layer (ALIAS Norm), ReLU, and Conv layer.
First, a resized hair-agnostic image $\mathbf{I}_{HA}^{i} \in \mathbb{R}^{3 \times H^{i} \times W^{i}}$, an RGB-rendered facial keypoint image $\mathbf{K}_{s}^{i} \in \mathbb{R}^{3 \times H^{i} \times W^{i}}$, and the warped target hair image $(\mathbf{I}_{t,w}^{h})^{i}$ are concatenated and fed to a Conv layer to obtain ${f}^{i}$.
Then, ${f}^{i}$ is concatenated with ${h}^{i}$, the feature from the previous layer, and injected to $i$-th ALIAS ResBlk.

\noindent\textbf{ALIAS Norm.} ALIAS Norm normalizes ${h}^{i} \oplus {f}^{i}$ separately based on the resized inpainting mask $\mathbf{M}_{inpaint}^{i} \in \mathbb{R}^{1 \times H^{i} \times W^{i}}$.
Then, the normalized features are de-normalized with affine parameters $\mathbb{\gamma}$ and $ \mathbb{\beta}$, estimated based on the resized $\mathbf{M}_{div}^{i}$. $\mathbf{M}_{div}^{i}$ consists of three components, the warped target hair mask $\mathbf{M}_{t,w}^{h,i}$, semantic-region-aware inpainting mask $\mathbf{M}_{SI}^i$, and  $\mathbf{M}_{inpaint}^i$.

\noindent\textbf{\maskblock.} The network separates $\mathbf{M}_{inpaint}$ into face, clothes, background, and unknown region.
\maskblock consists of two Down blocks, one Conv layer, and two Up blocks.
Each Down block has a Conv, Batch-Norm, and ReLU layer. Also, each Up block has an up-sampling layer, Conv, Batch-Norm, and ReLU layer. \Rthree{\maskblock is trained end-to-end with \hairsynthesis generator.}

\noindent\textbf{Lossses.} The details of the losses we use are described below.

\noindent\textit{$\mathcal{L}_{cGAN}$, $\mathcal{L}_{FM}$, and $\mathcal{L}_{percept}$.} We adopt the conditional adversarial loss $\mathcal{L}_{cGAN}$, the feature matching loss $\mathcal{L}_{FM}$, and the perceptual loss $\mathcal{L}_{percept}$, referring to VITON-HD, SPADE, and pix2pixHD~\cite{choi2021viton, park2019SPADE, wang2018high}. We use the hinge loss for $\mathcal{L}_{cGAN}$~\cite{zhang2018unreasonable}. 
Let $\mathrm{D}$ be the discriminator and $\mathrm{D}^{i}$ be the activation of the $i$-th layer $\mathrm{D}$. 
Similarly, $\mathrm{VGG}^{i}$ be the activation of the $i$-th layer \Rthree{VGG19} network~\cite{simonyan2014very}.
$N_{\mathrm{D}^i}$ and $N_{\mathrm{VGG}^i}$ are the number of elements in $\mathrm{D}^{i}$ and $\mathrm{VGG}^{i}$, respectively.
Each of the above loss functions is described below.
\vspace{-0.2cm}
\begin{equation}
    \mathcal{L}_{cGAN} = \mathbb{E}[\log(\mathrm{D}(\mathbf{M}_{div},\mathbf{I_s}))] + \mathbb{E}[1-\log(\mathrm{D}(\mathbf{M}_{div},\mathbf{\hat{I}}))]
\end{equation}
\vspace{-0.5cm}
\begin{equation}
    \mathcal{L}_{FM} = \mathbb{E}\sum_{i=1}^T \frac{1}{N_{\mathrm{D}^i}}\rVert \mathrm{D}^i(\mathbf{M}_{div}, \mathbf{I_s}) - \mathrm{D}^i(\mathbf{M}_{div}, \mathbf{\hat{I}}))\rVert_{1}
\end{equation}
\vspace{-0.3cm}
\begin{equation}
    \mathcal{L}_{percept} = \mathbb{E}\sum_{i=1}^V \frac{1}{N_{\mathrm{VGG}^i}}\rVert \mathrm{VGG}^i(\mathbf{I_s}) - \mathrm{VGG}^i(\mathbf{\hat{I}})\rVert_{1}
\end{equation}
\noindent\textit{Hairstyle loss $\mathcal{L}_{style}$.}
To capture the fine details of hairstyle features, we utilize the Gram matrix~\cite{gatys2016image}.
We compute the L2 distance between the gram matrices of the generated hair features and the target hair features extracted by $\mathrm{VGG}$16~\cite{simonyan2014very}. The generated hair features are obtained based on $\mathbf{M}_g^h \otimes \mathbf{\hat{I}}$ and the target hair features are obtained based on $\mathbf{M}_t^h \otimes \mathbf{I_t}$. 
$\mathrm{G}^{i}$ is the $i$-th Gram matrix, $\mathrm{G}^{i}(v^{i}) = {v^{i}}^{\intercal}v^{i}$, where $v^{i} \in  \mathbb{R}^{H^{i}W^{i} \times N_{C^{i}}} $ is the activation of the $i$-th layer of $\mathrm{VGG}$.
Here, $N_{C^{i}}$ and $N_{\mathrm{G}^i}$ represent the number of channels in $\mathrm{VGG}^{i}$ and in $\mathrm{G}^{i}$, respectively. % N_{C^i} x N_{C^i}
The activations from $\{relu1\_2, relu2\_2, relu3\_3,relu4\_3\}$ of $\mathrm{VGG}$ are used for the loss.
\vspace{-0.2cm}
\begin{equation}
    \mathcal{L}_{style} = \mathbb{E}\sum_{i=1}^V \frac{1}{N_{\mathrm{G}^i}}\rVert \mathrm{G}^i(\mathrm{VGG}^i(\mathbf{M}_t^h \odot \mathbf{I_t})) - \mathrm{G}^i(\mathrm{VGG}^i(\mathbf{M}_g^h \odot \mathbf{\hat{I}}))\rVert_{2}
\end{equation}

\noindent\textit{SIM estimator loss $\mathcal{L}_{SIM}$.} $\mathbf{GT}_{SIM}$ is a ground truth segmentation mask of the inpainting mask $\mathbf{M}_{inpaint}$. $\mathbf{GT}_{SIM}$ is obtained from the source semantic masks of a face, clothes, and backgrounds extracted by the pre-trained face-parsing network~\cite{yu2018bisenet}. 
We compute the binary cross-entropy loss between $\mathbf{GT}_{SIM}$ and the predicted $\mathbf{M}_{SI}$ as below.
\vspace{-0.1cm}
\begin{equation} % check the equation BCE
    \mathcal{L}_{SIM} = -\mathbb{E}[\mathbf{GT}_{SIM}\log(\mathbf{M}_{SI}) + (1-\mathbf{GT}_{SIM})\log(1-\mathbf{M}_{SI})]     
\end{equation}
\noindent The total loss of the hair synthesis module is calculated as follows:
\vspace{-0.2cm}
\begin{equation}
    \mathcal{L}_{total} = \mathcal{L}_{cGAN} + \lambda_{FM} \mathcal{L}_{FM} +
    \lambda_{percept} \mathcal{L}_{percept} + \lambda_{style} \mathcal{L}_{style} + \lambda_{SIM} \mathcal{L}_{SIM},
\end{equation} 
where we set both $\lambda_{FM}$ and $\lambda_{percept}$ to 10, $\lambda_{style}$ to 50, and $\lambda_{SIM}$ to 100.

% hyper parameters, data preprocessing details?
% 모델 구조 디테일, implementation details

We adopt the Adam optimizer~\cite{kingma2014adam} with $\beta_1 = 0$, $\beta_2 = 0.9$. The learning rate of the generator and the discriminator are set to 0.0001 and 0.0004, respectively. \Rthree{We train the hair synthesis module for 15,000 iterations with batch size 8 for both K-hairstyle and VoxCeleb.}

\begin{figure*}[h!]
    \centering
    \includegraphics[width=\linewidth]{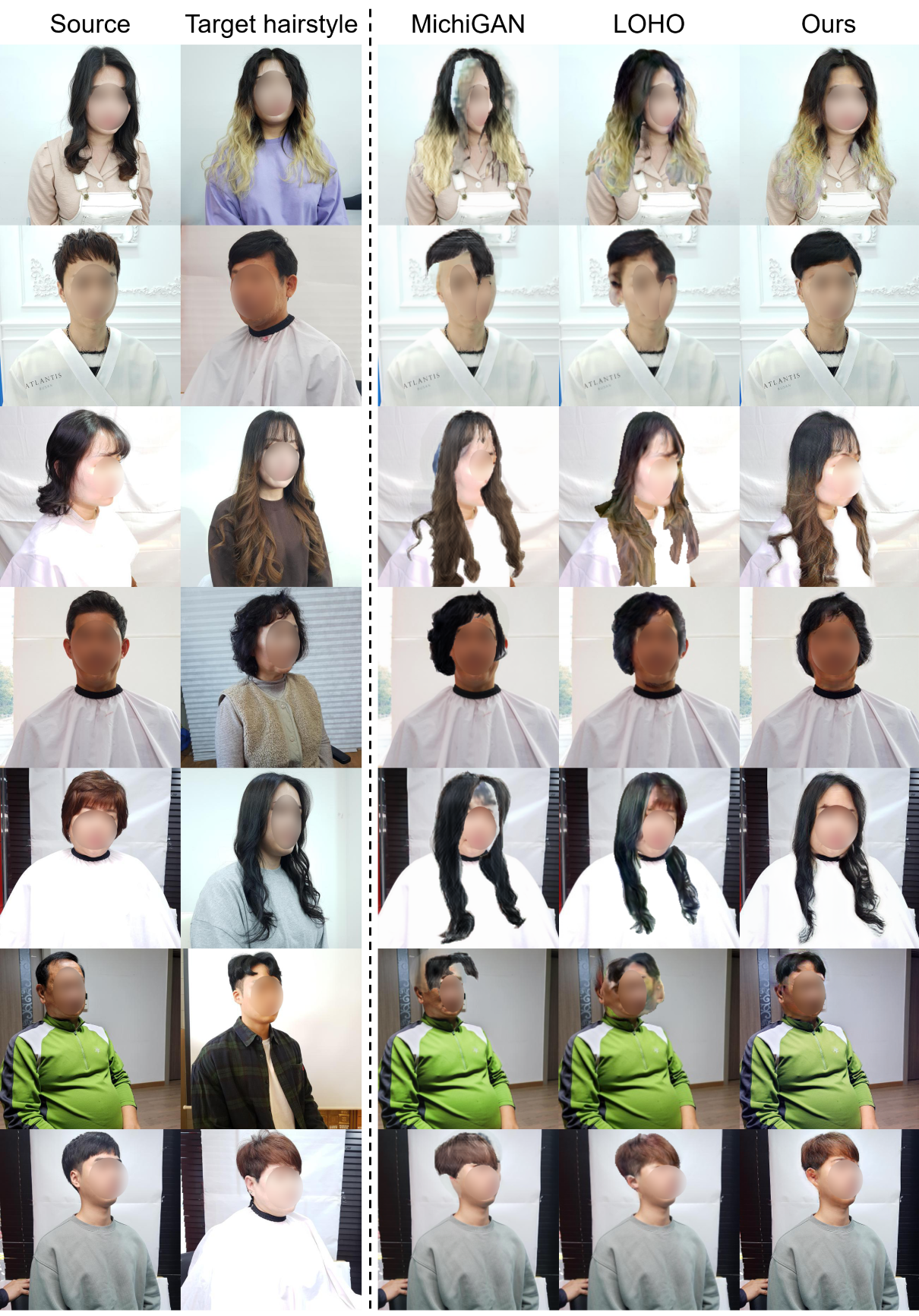}
    \vspace{-0.8cm}
    \caption{Qualitative comparison with K-hairstyle dataset. Due to the privacy issue, we blur the faces of the images.}
%    \vspace{-0.2cm}
    \label{fig:qualitative_k} 
\end{figure*} 

\begin{figure*}[h!]
    \centering
    \includegraphics[width=\linewidth]{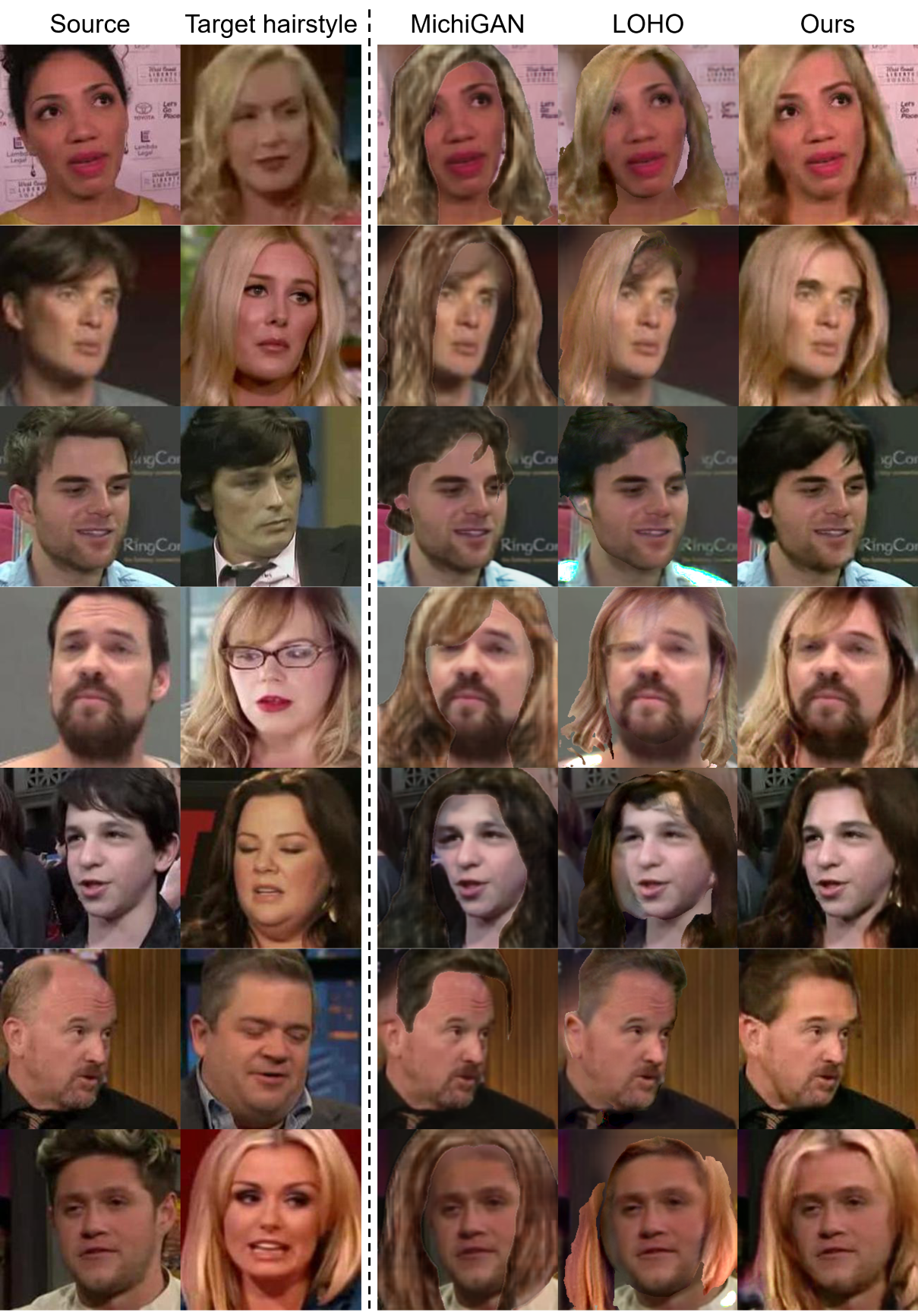}
    \vspace{-0.8cm}
    \caption{Qualitative comparison with VoxCeleb dataset.}
%    \vspace{-0.2cm}
    \label{fig:qualitative_v} 
\end{figure*}

\vspace{-0.3cm}
\section{Additional Qualitative Results}
\vspace{-0.2cm} 
We conduct an additional qualitative comparison between our model and the baseline models. Fig.~\ref{fig:qualitative_k} and Fig.~\ref{fig:qualitative_v} present qualitative results of K-hairstyle dataset and VoxCeleb dataset, respectively.
The results show that \ourmodel successfully transfers hairstyles even when the source and the target image have different poses.
Furthermore, our model preserves delicate target hairstyle features (\eg curl, two-toned hair color, etc.) better than other models.

\section{Limitations}
\begin{figure}[h!]
    \centering
    \includegraphics[width=0.7\linewidth]{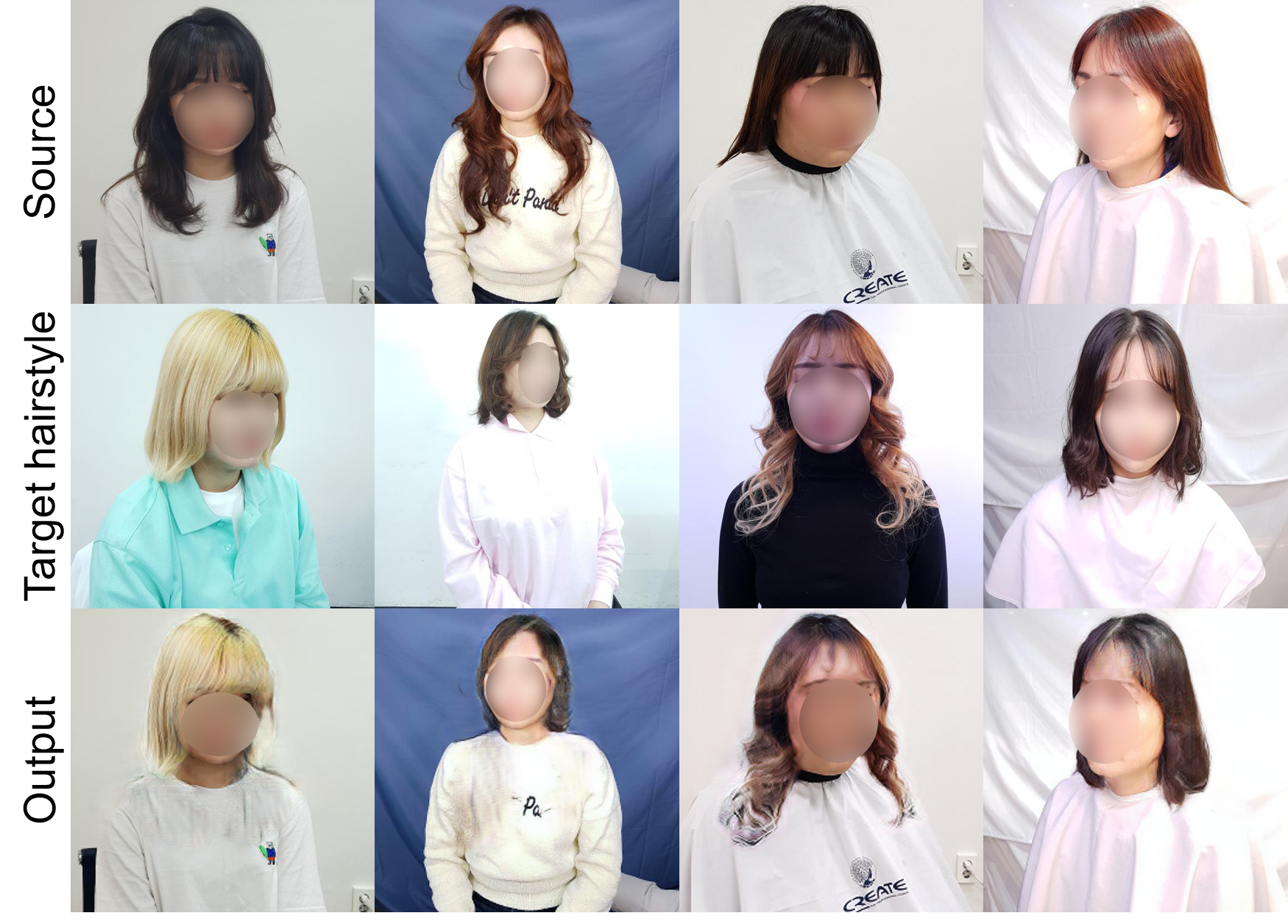}
    \caption{Limitations of \ourmodel. The first column describes the case where a target hair has an extreme occlusion and the second column is an example of complicated occlusion regions in a source image. The last two columns are the cases of limitation due to inaccurate hair segmentation masks.}
    \vspace{-0.3cm}
    \label{fig:limitation} 
\end{figure}

Although \ourmodel successfully transfers a hairstyle between images of different poses, our model has several limitations.
First, \ourmodel has difficulty in aligning a target hair which has a significant occlusion of hair. Since our hair alignment module utilizes a warping operation to align the target hair, the module can rearrange the pixels of the existing hair but cannot newly generate the unseen hair. The first column of Fig.~\ref{fig:limitation} presents an example where the right side of the target hair is extremely occluded. In this case, our model cannot transfer the right hair of the target image to the source image. 

Next, a complicated texture or structure in occlusion regions degrades the quality of generated images. For instance, as described in the second column of Fig.~\ref{fig:limitation}, even though a person in a source image wears clothes with complex patterns, our model inpaints the region only with simple and general texture.

Lastly, \ourmodel is dependent on hair segmentation masks. For example, if the target hair mask contains irrelevant regions such as the forehead, the output inevitably contains the region. On the third column of Fig.~\ref{fig:limitation}, the forehead of the output reflects the target forehead color which is different from the source since the target hair mask includes its forehead.
Also, the last column of Fig.~\ref{fig:limitation} indicates an example where the target hair mask does not contain thin hair on the forehead, leading to an inaccurate hairstyle transfer.

\begin{table}[h!]
    \centering
    \small
    \begin{tabular}{c|c|c}
    \toprule
    Dataset & K-hairstyle & VoxCeleb \\ 
    \midrule
    StarGAN v2 (Gender) & 0.5238 & 0.9938  \\ %\hline
    StarGAN v2 (Hair) & 0.5590 & - \\ %\hline
    \ourmodel (Ours) & \textbf{0.9892} & \textbf{0.9993}\\ 
    \bottomrule
    \end{tabular}%
    \vspace{0.2cm}
    \caption{Quantitative comparison of identity preservation performance between StarGAN v2 and \ourmodel. We measure the face verification accuracy using the pretrained ArcFace with K-hairstyle and VoxCeleb dataset.}
    \label{Table:stargan}
\end{table}

\section{Comparison with StarGAN v2}
%ICIP 논문을 보면 StarGAN이 엄청 잘한다 
\Rfour{The previous work~\cite{kim2021k} shows that StarGAN v2~\cite{choi2020starganv2} has a capability to modify the hairstyle of the source image based on the given target image.
However, we found that StarGAN v2 fails to preserve other features (\eg skin color, face shape, clothes, etc.) of the source image, which are essential to maintain the source person's identity.
% Since the hairstyle transfer requires modifying only the hairstyle, while preserving the identity of the source image, StarGAN v2 is inappropriate for our baseline. 
Since StarGAN v2 changes both the hairstyle and the identity of the source image, it is not appropriate to compare StarGAN v2 with our model.
For this reason, none of the existing hairstyle transfer work such as MichiGAN, LOHO considered StarGAN v2 as their baseline, either.}

\begin{figure}[h]
    \centering
    \includegraphics[width=\linewidth]{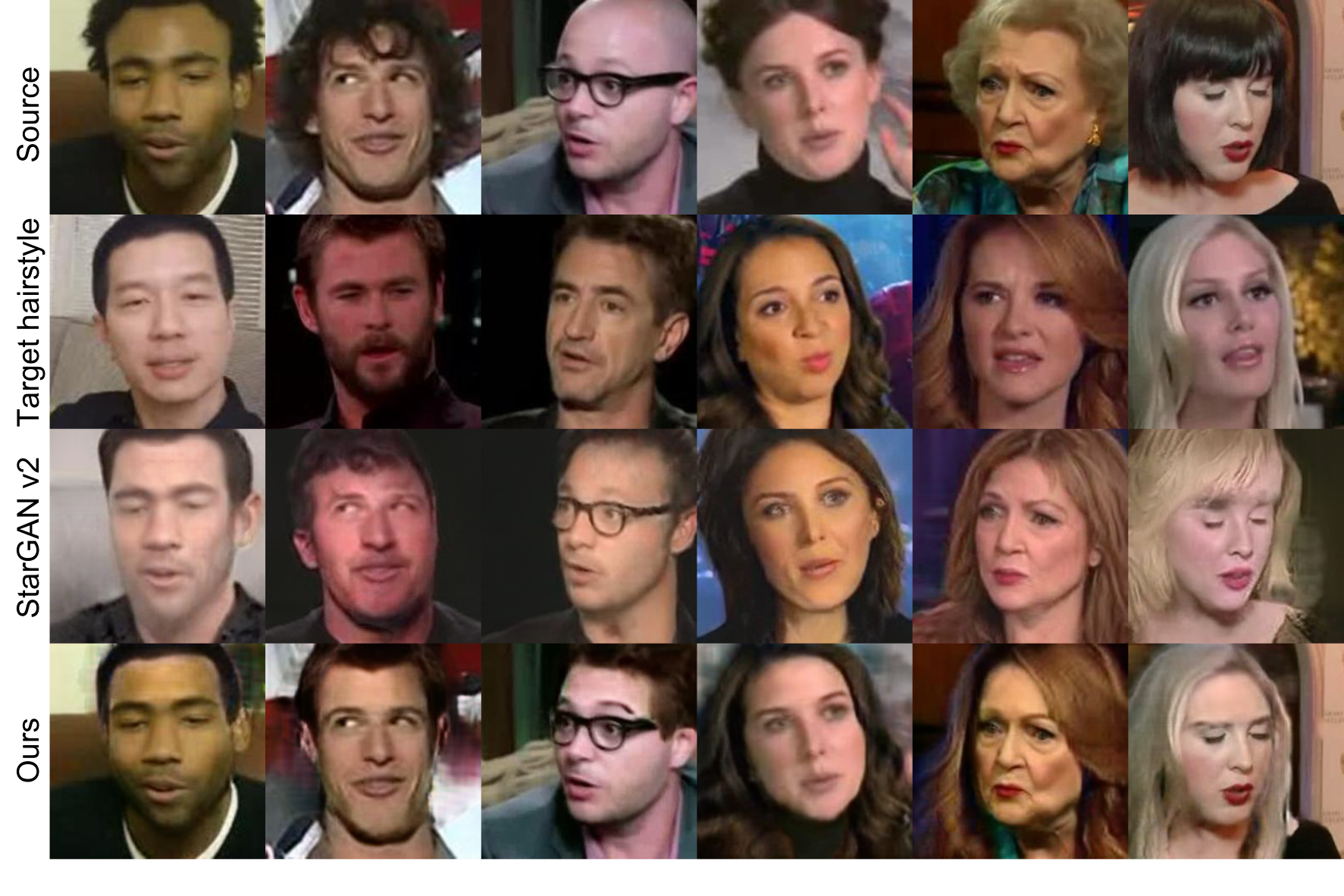}
    \vspace{-0.8cm}
    \caption{Qualitative comparison of identity preservation performance between StarGAN v2 and \ourmodel.}
    \vspace{-0.2cm}
    \label{fig:stargan} 
\end{figure}

\REV{To evaluate the identity preservation performance, we measure the face verification accuracy of StarGAN v2 compared to HairFIT using the pre-trained ArcFace~\cite{deng2018arcface}, which is one of the state-of-the-art face recognition models. 
Since StarGAN v2 requires the domain labels for training, we utilize the gender labels and the hairstyle labels provided from K-hairstyle dataset and the gender labels from VoxCeleb dataset.
% We first trained StarGAN v2 with both K-hairstyle and VoxCeleb datasets based on gender domains~\cite{kim2021k}. 
% We also trained StarGAN v2 with hairstyle domains since the K-hairstyle contains 31 hairstyle class labels, unlike VoxCeleb. 
As shown in Table~\ref{Table:stargan}, HairFIT successfully preserves the source identity in both K-hairstyle and VoxCeleb.
On the other hand, StarGAN v2 trained with K-hairstyle fails to preserve the source identity. 
% In contrast, StarGAN v2 trained with gender labels and hairstyle labels achieve 0.52 ad 0.55, respectively. 
% The result proves that HairFIT achieves higher accuracy with a large margin.
Although StarGAN v2 trained with VoxCeleb achieves the high verification accuracy, Fig.~\ref{fig:stargan} demonstrates that StarGAN v2 also modifies the features (\eg skin color, makeup style, etc.) related to the source identity.
This reason makes StarGAN v2 hardly applicable for hairstyle transfer.}

%\vspace{-0.6cm}

\end{document}